\documentclass{article}
\usepackage{wrapfig}
\usepackage{caption}
\usepackage{natbib}
\usepackage{times}
\usepackage[utf8]{inputenc} 
\usepackage[T1]{fontenc}    
\usepackage{booktabs}       
\usepackage{amsfonts}       
\usepackage{nth}
\usepackage{subfigure}
\usepackage{graphicx}
\usepackage{amsmath}
\usepackage{caption} 
\renewcommand{\vec}[1]{\mathbf{#1}}

\title{Siamese Encoding and Alignment by Multiscale Learning with Self-Supervision}

\author{
  Eric Mitchell \qquad Stefan Keselj\\
  Sergiy Popovych \qquad Davit Buniatyan \qquad H. Sebastian Seung\\
  \qquad\\
  Neuroscience Institute and Computer Science Department\\
  Princeton University
}

\begin{document}

\maketitle

\begin{abstract}
We propose a method of aligning a source image to a target image,
where the transform is specified by a dense vector field. The two
images are encoded as feature hierarchies by siamese convolutional nets. Then a hierarchy of aligner modules computes the transform in a coarse-to-fine recursion. Each module receives as input the transform that was computed by the module at the level above, aligns the source and target encodings at the same level of the hierarchy, and then computes an improved approximation to the transform using a convolutional net. The entire architecture of encoder and aligner nets is trained in a self-supervised manner to minimize the squared error between source and target remaining after alignment. We show that siamese encoding enables more accurate
alignment than the image pyramids of SPyNet, a previous deep learning
approach to coarse-to-fine alignment. Furthermore, self-supervision applies even without target values for the transform, unlike the strongly supervised SPyNet. We also show that our approach outperforms one-shot approaches to alignment, because the fine pathways in the latter approach may fail to contribute to
alignment accuracy when displacements are large. As shown by previous one-shot approaches, good results from self-supervised learning require that the loss function additionally penalize non-smooth transforms. We demonstrate that
"masking out" the penalty function near discontinuities leads to
correct recovery of non-smooth transforms. Our claims are supported by empirical comparisons using images from serial section electron microscopy of brain tissue.\footnote{Since this paper was originally submitted for publication in 2018, many of its general ideas have been published by others (e.g. \citet{hui2018liteflownet}). We are releasing the paper to arXiv in its original form as the specific application to serial section EM images is still novel and informative.}
\end{abstract}

\section{Introduction}

The computation of correspondences between the pixels of two images is a classic problem in computer vision, and occurs in diverse applications such as optical flow, stereo matching, and image alignment. Recently a number of deep learning approaches to dense image correspondences have been proposed. SPyNet starts from two image pyramids, and traverses the pyramid levels in a coarse-to-fine direction. The optical flow computed at the level above is used to warp one image to the other at the current level, before computing the residual optical flow \citep{ranjan2017optical}. If every level does its job correctly, the residual optical flow is always small, and therefore can be computed easily. While the recursive approach of SPyNet is intuitively appealing, it has been more popular to compute correspondences in one shot by a single convolutional net, as in FlowNet \citep{flownet} and other proposed algorithms \citep{WarpNet, luo2016efficient, yoo2017ssemnet, Unflow, balakrishnan2018unsupervised}. 

SPyNet has two significant weaknesses. First, as its creators note, ``it is well known that large motions of small or thin objects are difficult to capture with a pyramid representation.'' This is because many rounds of downsampling may deprive the coarse levels of the pyramid of information that is important for establishing correspondences. If the recursive computation makes an error at a coarse level, it will likely be impossible to recover from this error at the fine levels. This difficulty becomes more severe for taller pyramids with many levels. The second weakness is that SPyNet learning is strongly supervised, depending on the existence of ground truth values for the optical flow. This may be fine for optical flow, for which many ground truth datasets have been constructed, but limits the applicability of SPyNet to other image correspondence problems for which ground truth data may not currently exist. 

Here we address both weaknesses of SPyNet with a new method called Siamese Encoding and Alignment by Multiscale Learning with Self-Supervision (SEAMLeSS).  To align a source image to a target image, both images are encoded by siamese convolutional nets as hierarchies of feature maps. Then a sequence of convolutional nets computes the alignment transform in a coarse-to-fine recursion. The entire architecture of encoder and aligner nets is trained to minimize the squared error between source and target remaining after alignment. This self-supervised learning removes the need for ground truth correspondences. 

To compare SEAMLeSS with other algorithms, we use neuronal images from serial section electron microscopy (EM). The alignment of 2D images to create a 3D image stack is an important first step in reconstruction of the connectome. Because each 2D image is of a physically distinct slice of brain, image content differs across slices. Furthermore, the deformations of image slices cannot be well-approximated by affine transforms or other simple parametric classes of transforms.

If we replace the siamese encodings of SEAMLeSS with image pyramids, we obtain a self-supervised variant of SPyNet. We show that SEAMLeSS is more accurate, demonstrating that the siamese convolutional encodings provide more reliable information for alignment.

We also compare with FlowNet \citep{flownet}, which is representative of one-shot approaches to image correspondences \citep{balakrishnan2018unsupervised, yoo2017ssemnet}.  We show that self-supervised FlowNet is mostly inferior to SPyNet, except that the worst errors of SPyNet are more severe than the worst errors of FlowNet. This is consistent with the hypothesis that the worst errors of SPyNet likely originate at the coarse levels of the pyramid, which are information-poor due to downsampling. SEAMLeSS outperforms FlowNet as well as SpyNet, in both best and worst cases. This provides evidence that the recursive approach, if applied to learned hierarchical encodings rather than image pyramids, is superior to the one-shot approach.

The one-shot approaches \citep{yoo2017ssemnet, balakrishnan2018unsupervised} already introduced self-supervised learning and showed that good results require training with a loss function that includes a penalty for non-smooth transformations. However, such a penalty function is obviously problematic if the correct transformation contains singularities. For example, the optical flow is typically discontinuous at the boundaries of a foreground object that is moving relative to the background. To properly treat such situations, we propose that the penalty function be "masked out" at singularities. We apply this approach to serial section EM images, and show that it is able to correct discontinuous distortions at cracks. 

SEAMLeSS can be regarded as training a neural net to approximate the solution of an optimization problem, finding a smooth alignment transform that minimizes squared error. One could instead solve the optimization problem at inference time for each pair of images to be aligned. We show that this approach is one or two orders of magnitude slower than the neural net.


\section{Methods}
Given a source image $S$ and target image $T$, our goal is to find a transform $\vec{F}:R^2\to R^2$ that minimizes the squared error
$\sum_\vec{r}[(S\circ\vec{F})(\vec{r})-T(\vec{r})]^2$    
subject to regularization or constraints on $\vec{F}$. One approach in computer vision is to perform this optimization by some iterative method for each source/target pair; we quantitatively consider with this traditional approach in Sec. \ref{cracks}. Here we propose instead to train a neural network that takes source and target images as input, and outputs a transform $\vec{F}$ that approximately minimizes the squared error. The neural net is trained to minimize the squared error on average for a set of example source/target pairs. The neural net has the potential advantage of speed; approximating the optimal $\vec{F}$ might be faster than iterative optimization of the squared error. Below we explain our coarse-to-fine network architecture in which the transform $\vec{F}$ is computed recursively, based on a hierarchy of image encodings at different spatial resolutions.

\subsection{Siamese encoder nets}
\begin{figure}
	\centering
	\includegraphics[width=\textwidth]{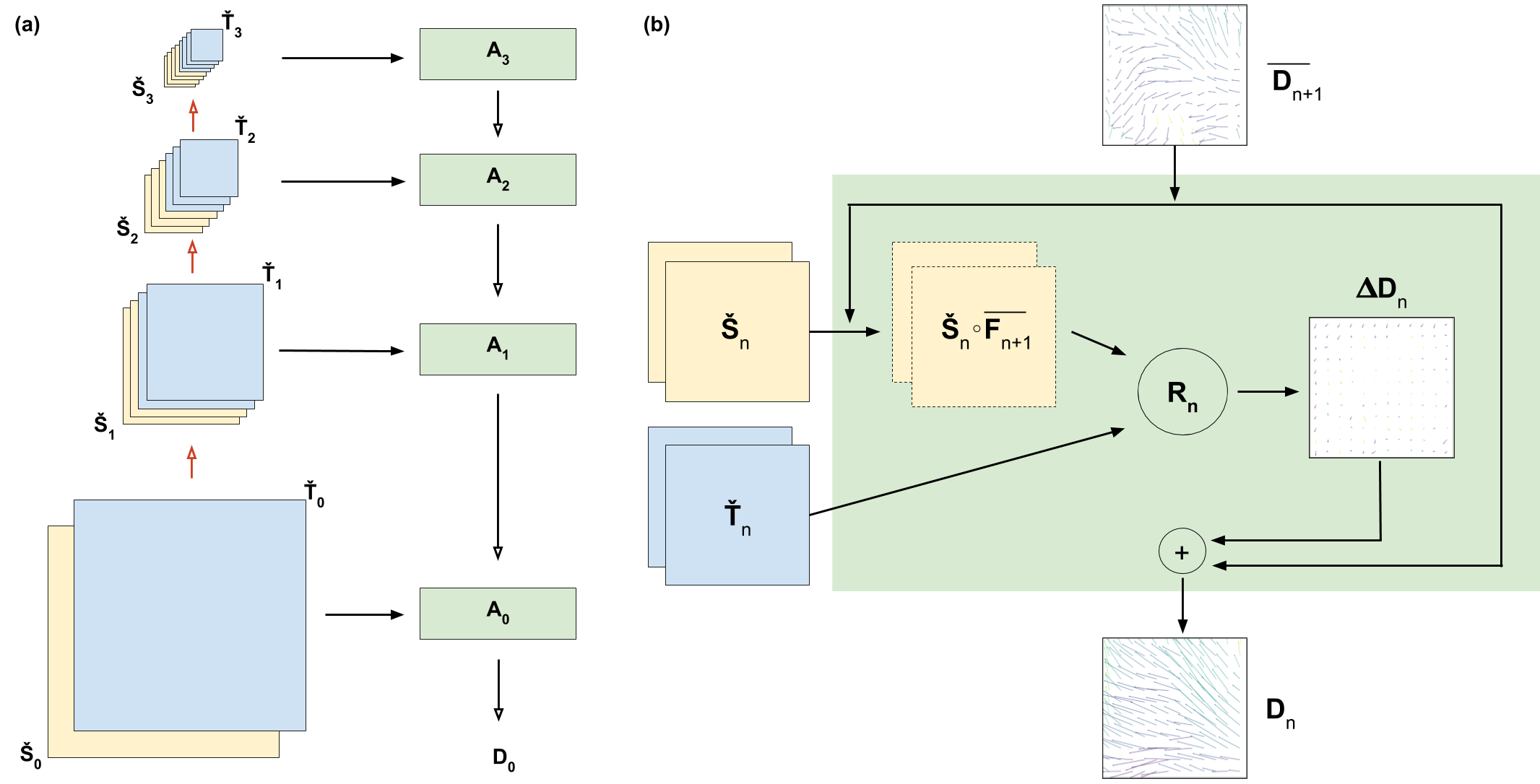}
	\caption[Architecture]{(Architecture a) Hierarchical encoding (left) and recursive coarse-to-fine alignment (right). Source and target images $S_0$ and $T_0$ are encoded by siamese convolutional nets. Each red arrow represents two $3\times 3$ convolutional layers followed by $2\times 2$ max pooling. MIP$n$ encodings $\text{\v{S}}_n$ and $\text{\v{T}}_n$ consist of $6n+6$ feature maps each (fewer are shown for simplicity). The MIP$n$ aligner module $A_n$ takes $\text{\v{S}}_n$ and $\text{\v{T}}_n$ as input, and the upsampled MIP$(n+1)$ displacement field, and generates as output the MIP$n$ displacement field. Each black arrow with hollow head is an upsampling. Our experiments had $N=7$ MIP levels, but $N=3$ are shown for simplicity. (Architecture b) Aligner module (right) for MIP$n$. The upsampled MIP$(n+1)$ transform $\overline{\vec{F}_{n+1}} = \rm{identity} + \overline{\vec{D}_{n+1}}$ is used to align the source encoding $\text{\v{S}}_n$ with bilinear interpolation, and the result is passed with $\text{\v{T}}_n$ to the residual aligner net $R_n$, which computes the residual displacement field $\Delta \vec{D}_n$. The latter is added to $\overline{\vec{D}_{n+1}}$ to yield the MIP$n$ estimate $\vec{D}_n=\Delta \vec{D}_n + \overline{\vec{D}_{n+1}}$ of the displacement. The aligner net has 5 layers of $7\times 7$ convolutions, with feature map counts of $(32, 64, 32, 16, 2)$ à la \citet{ranjan2017optical}.}\label{fig:architecture}
\end{figure}

Each level of the encoder network consists of convolutional layers followed by max pooling (Fig. \ref{fig:architecture}a, left). The set of feature maps after the $n$th max pooling will be called the MIP$n$ encoding of the image.  All the MIP levels together constitute a hierarchical encoding of the image containing multiple resolutions. 



The encoder network is trained using the same loss function as the aligner networks. Since the encoder networks for the source and target images share the same weights, they are said to be "siamese nets"\citep{chopra2005learning}. In this context, one can regard the aligner nets and the MSE loss as a similarity function for deep metric learning. 


\subsection{Aligner module}
We write the transform as $\vec{F}(\vec{r}) = \vec{r} + \vec{D}(\vec{r})$, where $\vec{D}$ is the displacement vector field. We likewise write $\vec{F}_n$ and $\vec{D}_n$ for the approximations computed at MIP level $n$.

Each aligner module $A_n$ is trained to output the MIP$n$ estimate $\vec{D}_n$ of the displacement, taking as input the upsampled MIP$(n+1)$ estimate $\overline{\vec{D}_{n+1}}$ of the displacement and the MIP$n$ encodings $\text{\v{S}}_n$ and $\text{\v{T}}_n$ of source and target images (Fig. \ref{fig:architecture}a, right).

Our aligner module is almost identical to that of SPyNet \citep{ranjan2017optical}, except that it takes encodings as input rather than images, and uses leaky ReLU \citep{maas2013rectifier} rather than ReLU. Another difference is that the output of the convolutional net in the aligner module is multiplied by 0.1, because displacements are typically smaller than the size of the image, and the coordinate system for the image runs from $[-1,-1]$ at the top-left corner and $[1,1]$ at the bottom-right corner. 

\subsection{Hierarchical training}
Training is done in stages, starting with the highest MIP level $N$ and ending with MIP level 0. The loss function for the $n$th stage of training is
\begin{equation}\label{eq:LossFunctionStage}
\sum_\vec{r}[(S_n \circ \vec{F}_n)(\vec{r})-T_n]^2 + \lambda \sum_{\langle\vec{r},\vec{r}'\rangle}\left|\vec{D}_n(\vec{r})-\vec{D}_n(\vec{r}')\right|^2
\end{equation}
where $S_n$ and $T_n$ are the source and target images downsampled to MIP$n$. The first sum is over pixel locations $\vec{r}$, and $(S_n \circ \vec{F}_n)(\vec{r})$ is estimated by bilinear interpolation. The second sum penalizes non-smooth displacement fields, and is over location pairs $\langle\vec{r},\vec{r}'\rangle$ that are separated by two pixels horizontally or two pixels vertically. This is the centered first-order finite difference approximation to the gradient of the displacement. The hyperparameter $\lambda$ controls the strength of the penalty term. We use $\lambda=0.1$ in our experiments.

Each stage of training consists of $k$ epochs, where a single epoch consists of a complete pass over every sample in our training set. In the $n$th stage, only $A_n$ is trained, except for the last epoch when all of $A_n,\ldots,A_N$ are trained. All weights of the encoder net are always being trained during every epoch of every stage. It was determined empirically that a value of 2 for $k$ yielded an effective tradeoff of fast training and good performance.

During training, the data is augmented by applying translation, rotation, scaling, brightness/contrast, defects, and cropping (see Appendix \ref{augmentation}) 

\subsection{Dataset}\label{dataset}
Our experiments are performed with images drawn from Pinky40, a dataset based on 1000 images acquired by serial section electron microscopy of mouse visual cortex. The imaged volume was millions of cubic microns. We downsampled all images by three rounds of $2\times 2$ average pooling, after which the pixel size was $32\times 32\times 40$ nm$^3$. 
Our experiments used a version of Pinky40 which had been "prealigned" using globally affine transforms only. Misalignments as large as $250$ pixels remain, because the image deformations are poorly approximated by global affine transforms.  For the most part, the remaining distortions in Pinky40-prealigned are smooth and locally affine (large translations and small rotations and scaling).  However, there are also discontinuities, such as cracks and folds (see Sec. \ref{cracks}).


For comparison, we also have a version of Pinky40 that was aligned using custom software package called Alembic, which is based on traditional computer vision methods following \citet{saalfeld2012elastic}. Alembic does block matching by normalized cross correlation to create an elastic spring mesh that models image correspondences and a prior on image deformations, and then relaxes the spring mesh using conjugate gradient. Human intervention was used to tune block matching parameters for individual slices if needed, and also to remove erroneous block matches. Pinky40-Alembic, by incorporating both hand-designed computer vision and human correction, provides a highly accurate benchmark against which to compare the methods of this paper.

\subsection{Quantification of alignment accuracy}
Our evaluation data consists of 50 subvolumes of Pinky40-prealigned, each containing 50 consecutive $1132 \times 1132$ slices. These subvolumes are distinct from those in the training set. Accuracy of alignment is quantified with a chunked Pearson correlation metric (CPC) \citep{garcia2007quality}. 

To evaluate an alignment system, we first run it on each of the test volumes by iteratively aligning slice $(k+1)$ to aligned slice $k$ for $0 \le k < 49$. Every slice of the resulting aligned image stack is then chunked into $12\times 12$ non-overlapping chunks of equal size. We compute the Pearson correlation coefficient (Pearson's $r$) between chunk $(i,j)$ in slice $k$ of stack $l$  and chunk $(i,j)$ in slice $(k+1)$ of stack $l$ for all $0\le  i,j<12$, $0 \le k < 49$, $0\le l < 50$. Pinky40 contains some regions where image data is missing. To prevent these regions from affecting the alignment score, we do not compute a correlation coefficient for any chunk that has any missing data (less than $4\%$ of all chunks are discarded).

This procedure yields roughly 140,000 valid Pearson correlation coefficients for each alignment system. The  distributions of these coefficients can then be compared between alignment systems. In the following sections we quantitatively compare SEAMLeSS with SPyNet \citep{ranjan2017optical} and FlowNet \citep{flownet}, two neural net architectures originally applied to optical flow and adapted here to our alignment problem. We also compare with Alembic as a baseline.


\section{Experiments}

We first compare SEAMLeSS and SPyNet. SPyNet was originally trained to compute optical flow using supervised learning \citep{ranjan2017optical}. Here we trained the same SPyNet architecture to align images using self-supervised learning with the loss function of Eq. (\ref{eq:LossFunctionStage}). When trained in this way, SPyNet only differs from SEAMLeSS in its use of image pyramids rather than convolutional net encodings. We also added one more MIP level to the SPyNet image pyramid to match our SEAMLeSS architecture. 

Alignment accuracy is quantified using the CPC distributions (Methods). 
Based on comparing the distribution means, SEAMLeSS has superior accuracy to SPyNet (Table~\ref{alignresultstable}). SEAMLeSS also comes out superior when we compare various percentile values of the distributions.

\begin{table}[htp]
	\centering
	\caption{Comparing distributions of chunked Pearson correlations}\label{alignresultstable}
	\begin{tabular}{lcccccc} \toprule[2px] 
		                      & \multicolumn{2}{c}{Summary} & \multicolumn{4}{c}{Percentile Values} \\ \cmidrule(r){2-3} \cmidrule(r){4-7}
		System            & $\mu$ & $\sigma^2$ & \nth{1} & \nth{5} & \nth{95} & \nth{99} \\ \cmidrule(r){1-1}  \cmidrule(r){2-7}
		None          & 0.224 & 0.0300 & -0.0585 & -0.0133 & 0.524 & 0.651  \\
		FlowNet        & 0.417 & 0.0232 &  0.105 & 0.164 & 0.636 & 0.754 \\
		SPyNet        & 0.497 & 0.0241 &  0.0421 & 0.172 & 0.686 & 0.900 \\
		Alembic       & 0.528 & 0.0107 & 0.231 & 0.358 & 0.667 & 0.866  \\
		SEAMLeSS      & 0.545 & 0.0140 & 0.148 & 0.343 & 0.693 & 0.912 \\
		\bottomrule
	\end{tabular}
\end{table}

\begin{figure}[htp]
    \centering
    \subfigure[CPC distributions for each neural network architecture we examined]
        {\includegraphics[width=0.44\textwidth]{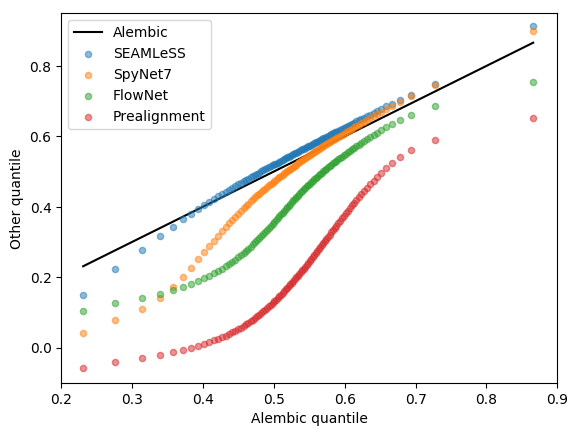}}
    \hfill
    \subfigure[CPC distributions for SEAMLeSS and SPyNet architectures using only the top 2 levels of the respective hierarchies]
        {\includegraphics[width=0.44\textwidth]{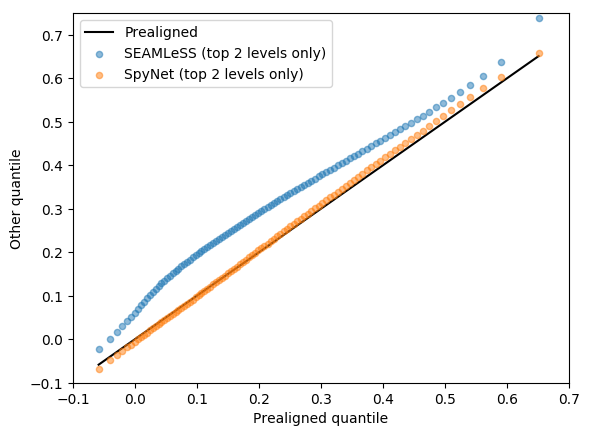}}
    \caption{Quantile-quantile (QQ) plot comparing the CPC distributions for various alignment systems. Confidence intervals were computed by bootstrapping, and are not shown as they are much smaller than the dots for the data.}
    \label{qqalign}

\end{figure}

To compare performance using the entire CPC distribution rather than summary statistics, the quantile-quantile (QQ) plot of Fig.\ref{qqalign} is helpful. In particular, the QQ plot highlights the differences in the tails of the CPC distributions. We see that SEAMLeSS and SPyNet are almost identical in the best case (right side of Fig. \ref{qqalign}). But SPyNet's worst errors tend to be substantially worse than those of SEAMLeSS (left side of Fig. \ref{qqalign}).

\begin{figure}[htp]
	\centering
	\subfigure[]
	{\centering\includegraphics[width=0.2\textwidth]{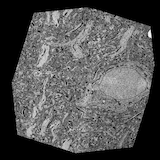}}
	\subfigure[]
	{\centering\includegraphics[width=0.2\textwidth]{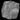}}
	\subfigure[]
	{\centering\includegraphics[width=0.2\textwidth]{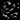}}
	\subfigure[]
	{\centering\includegraphics[width=0.2\textwidth]{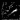}} \\
	\caption{Pixels vs. features (a) Original image (b) MIP6 of image pyramid has lost information by downsampling, which could lead to alignment errors. (c, d) Two MIP6 feature maps of SEAMLeSS provide complementary information, suggesting that the total of 48 MIP6 feature maps are rich in information for alignment}\label{fms}
\end{figure}

This empirical finding is consistent with our original hypothesis that the high MIP levels in the image pyramid may be information-poor, leading to errors by the aligner module. Fig. \ref{fms} is a visual comparison between the image pyramid and the convolutional encodings at a high MIP level. 

In our experiments, we found that SPyNet and SEAMLeSS have similar accuracy if there are only a few MIP levels in the hierarchy. The difference only becomes marked if there are many MIP levels. After many rounds of downsampling, the high MIP levels of the image pyramid are sparse in information. And if there is a large misalignment error at the top of the hierarchy, the subsequent coarse-to-fine recursion cannot correct it.

The SPyNet creators are aware of the limitations of the image pyramid, writing that "small or thin objects that move quickly effectively disappear at
coarse pyramid levels, making it impossible to capture their
motion" \citet{ranjan2017optical}.  Our results suggest that a convolutional encoder is the solution to this problem. Similarly, convolutional encoding has already replaced image pyramids in multiscale architectures for other problems in computer vision \citep{burt1987laplacian, yoo2015multi}.

\subsection{Comparing with one-shot approaches}
An alternative to the recursive coarse-to-fine approach is one-shot: a single convolutional net takes the full resolution source and target images as input, and generates a full resolution displacement field as output. No intermediate alignment of input images is involved in the computation. FlowNet \citep{flownet} was the first example of the one-shot approach, and has been followed by other similar papers \citep{balakrishnan2018unsupervised, yoo2017ssemnet, Unflow, WarpNet}. Like SPyNet, FlowNet was originally proposed as a supervised learning approach to optical flow. We retrained FlowNet to align images in a self-supervised manner using Eq. (\ref{eq:LossFunctionStage}) at MIP0 only. 

According to Fig. \ref{qqalign}, SPyNet is typically more accurate than FlowNet, but SPyNet's worst errors are typically worse than FlowNet's. In other words, SPyNet does not strictly dominate FlowNet. On the other hand, SEAMLeSS dominates both SPyNet and FlowNet at all quantiles. 

\subsection{Non-smooth deformations}
Self-supervised learning is convenient, and contrasts with the supervised learning of the original SPyNet and FlowNet papers \citep{ranjan2017optical,flownet}, which required target values for the transformation. However, good results require that the loss function for training include a term that penalizes non-smooth deformations. This was already observed by papers on the one-shot approach to alignment \citep{balakrishnan2018unsupervised,flownet,yoo2017ssemnet}. A smoothness prior is potentially problematic in many image correspondence problems. For example, optical flow is nonsmooth at the boundaries of a foreground object moving relative to the background. Depth is also discontinuous at the boundaries of objects in stereo matching.

Most of the previous papers applied self-supervised learning to image alignment, where the assumption of smoothness is often reasonable. But it is important to devise methods of dealing with discontinuities, if we want to extend the applicability of self-supervised learning to image correspondence problems more generally.

\begin{figure}[htp]
	\centering
	\subfigure[]
	{\centering\includegraphics[width=0.2\textwidth]{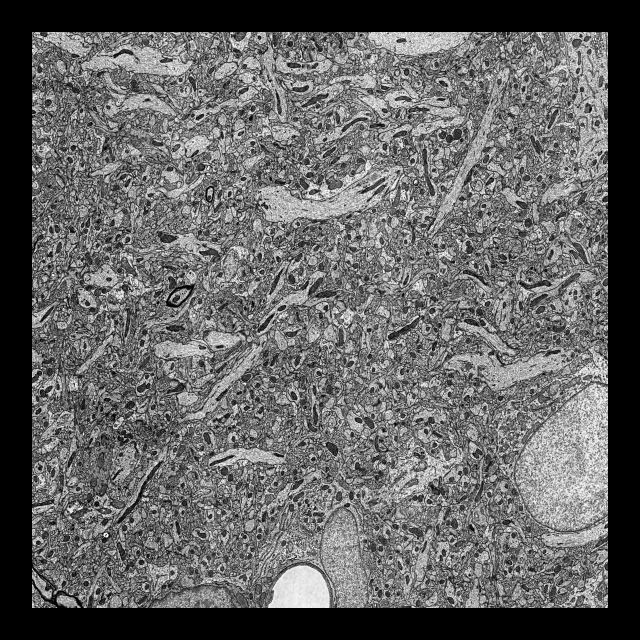}}
	\subfigure[]
	{\centering\includegraphics[width=0.2\textwidth]{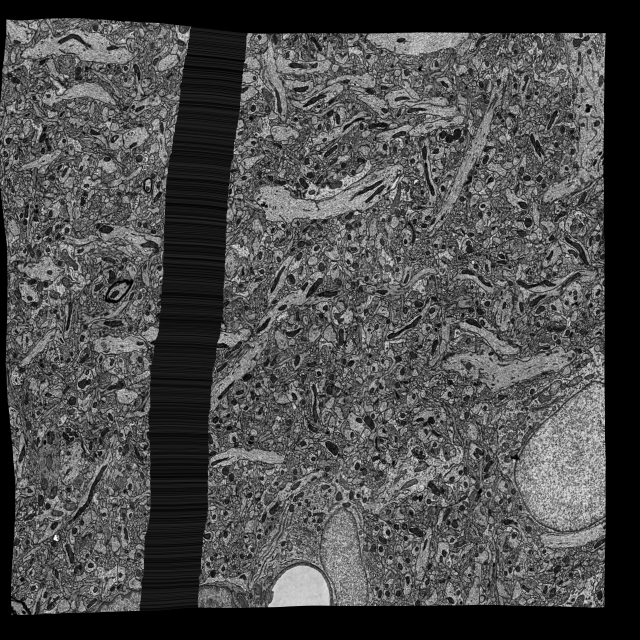}}
	\subfigure[]
	{\centering\includegraphics[width=0.2\textwidth]{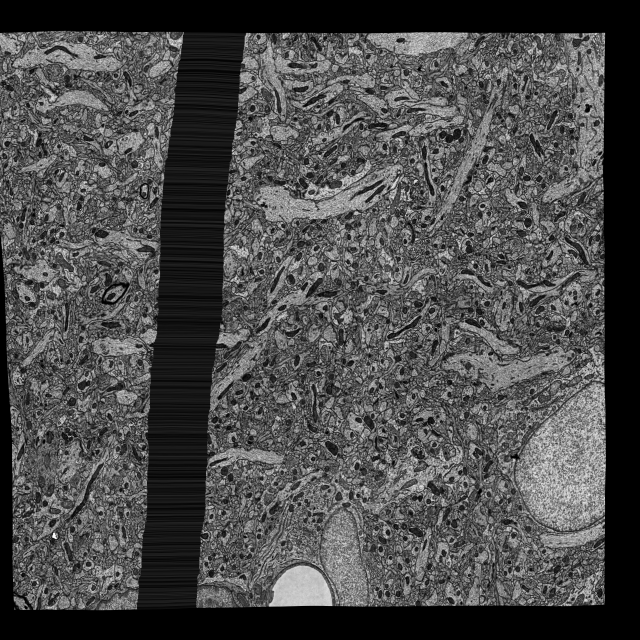}}
	\subfigure[]
	{\centering\includegraphics[width=0.2\textwidth]{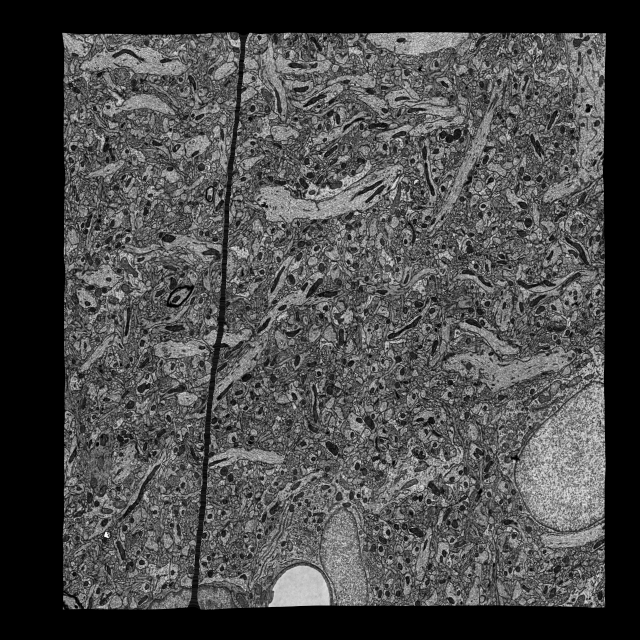}} \\
	\caption{How SEAMLeSS can learn to correct non-smooth deformations. (a) Target image, (b) Source image aligned by net that was not trained on cracks, (c) Source image is no better if aligned by net that was trained on cracks. (d) Source image is much better if aligned by net that was trained on cracks after "masking out" the loss function penalty on nonsmooth transforms near the cracks. The crack is almost completely ``closed'' by the net so that the aligned source matches the target.}\label{crackmaskcompare}
\end{figure}

Interestingly, images from serial section electron microscopy are not only continuously deformed, but can also be corrupted by discontinuous transformations. During the collection process, a serial section may physically crack or fold. On either side of a crack, the tissue is displaced outward, leaving a long stripe of white pixels. On either side of a fold, the tissue is displaced inward, leaving a narrow stripe of black pixels where the section is folded over itself. Examples of real cracks and folds can be found in the Appendix (Figs.~\ref{cracks} and~\ref{folds}). We developed a method of augmenting our training data by creating artificial cracks in images (Appendix), and used it for the experiments in Fig. \ref{crackmaskcompare}.


A SEAMLeSS network is unable to correct a wide crack, if it is trained without ever seeing cracks (Fig.~\ref{crackmaskcompare}b).  Even if the net is trained so that half of all examples contain cracks, it is still unable to correct a wide crack (Fig.~\ref{crackmaskcompare}c). Finally, if the net is trained with cracks in half of training examples and the penalty for nonsmooth displacements in Eq.~(\ref{eq:LossFunctionStage}) is eliminated at the crack pixels, then the net is able to correct the crack almost completely (Fig.~\ref{crackmaskcompare}c). Similar results are seen with real cracks (not shown). 

\begin{figure}[htp] 
    \centering
    \subfigure[]{%
        \centering\includegraphics[width=0.33\textwidth]{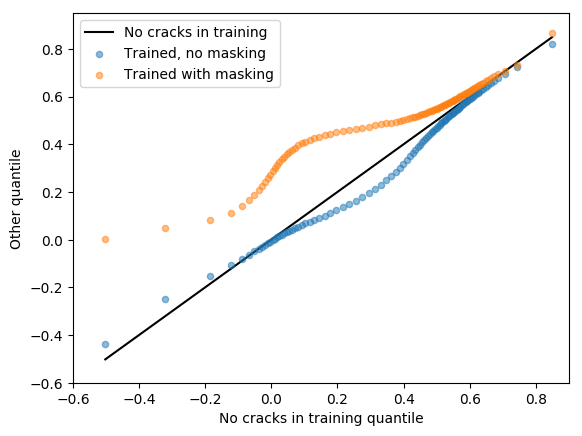}
        \label{fig:crackqq}
        }%
    \hfill%
    \subfigure[]{%
        \centering\includegraphics[width=0.33\textwidth]{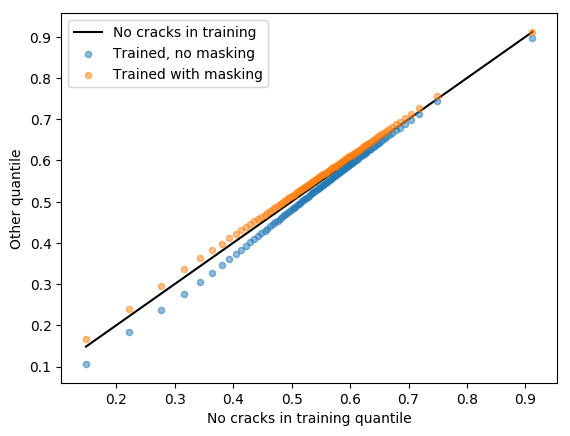}%
        \label{fig:qqcracktrainnnocrack}%
        }%
    \hfill%
    \subfigure[]{%
        \centering\includegraphics[width=0.33\textwidth]{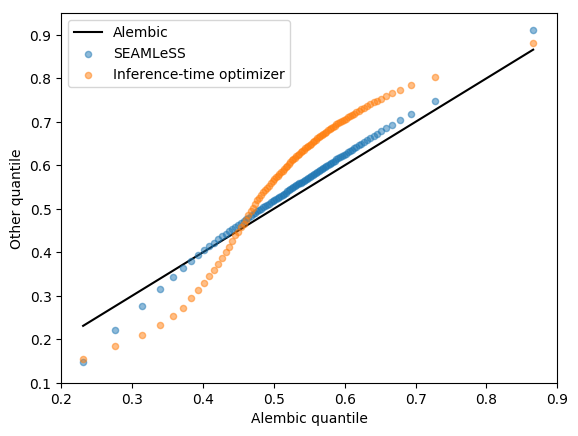}%
        \label{fig:qqopt}%
        }%
    \caption{More QQ plots corresponding to further experiments. We compare the performance of networks trained with no cracks added to the training set, cracks added to the training set but no masking, and cracks added to the training set with masking. (a) shows the performance of each training paradigm on a test set with cracks added to every slice and (b) shows the performance of the same networks with no cracks added to the test set. (c) shows the binary nature of inference-time optimization.}
\end{figure}

The QQ plot of Fig.~\ref{fig:crackqq} quantifies the resulting accuracy on a test set containing examples with and without cracks.
The masking of the nonsmoothness penalty during training gives a significant accuracy boost. Surprisingly, training on cracks without masking yields a net that is less accurate than a net that never saw cracks during training. Perhaps training the net without masking hurts accuracy in examples without cracks.

Our training with cracks is still self-supervised; since the cracks are artificially generated we know exactly where to mask out the nonsmoothness penalty. This method could be extended to discontinuities that occur naturally in the training set, rather than being artificially synthesized. If one has an automated means of locating candidate discontinuities in the training set, that could be used to mask out the nonsmoothness penalty. For example, object boundaries could be detected by a convolutional net in optical flow and stereo matching applications.

\subsection{Optimization at inference time}
SEAMLeSS computes an alignment of source to target image, and is trained to minimize the sum of squared error and a penalty for nonsmoothness in Eq. (\ref{eq:LossFunctionStage}), averaged over the examples in the training set. Alternatively, we could dispense with learning altogether and solve the optimization problem in Eq. (\ref{eq:LossFunctionStage}) at inference time for any source/target pair. This is done by regarding the vector field $\vec{F}$ as variables to be optimized, rather than generated by a neural network unlike in approaches \citep{weinzaepfel2013deepflow, revaud2015epicflow}. 

Direct gradient-based optimization of Eq. (\ref{eq:LossFunctionStage}) tends to get stuck in local minima if done only at the lowest MIP level. Therefore we adopt a coarse-to-fine hierarchical approach, much like the SEAMLeSS architecture. We optimize Eq. (\ref{eq:LossFunctionStage}) at the highest MIP level starting from random initial conditions. We upsample the resulting vector field and use it to initialize Eq. (\ref{eq:LossFunctionStage}) at the next highest MIP level, and so on. The gradient step size hyperparameter is increased as the MIP level decreases. Both this and the hyperparameter $\lambda$ controlling the nonsmoothness penalty in Eq. (\ref{eq:LossFunctionStage}) can be tricky to adjust. We stop early at each level, and only iterate to complete convergence at MIP0. In total, up to 10,000 iterations are required.


\section{Discussion} 


\textbf{Learned features are better than pixels. } Fig.\ref{fms} highlights the problems with using raw downsampling to acquire global context; Fig.\ref{qqalign}(b) provides empirical support for the idea that the image pyramid fails to produce meaningful results at high enough MIP levels as well as the claim that using learned feature maps can improve performance significantly.



\textbf {Coarse-to-fine can strictly dominate one-shot. } As seen in Fig.\ref{qqalign} and Table~\ref{alignresultstable}, one-shot alignment approaches (represented by FlowNet) lack in performance as compared to hierarchical approaches. FlowNet can be viewed as a modification of UNet, a well-known multiscale architecture \citep{UNet}. The data suggests that the fine scale pathways of FlowNet are not contributing much to the alignment. This is likely because the receptive fields of the neurons in the fine scale pathways are too small to contribute when displacements are large. The problem persists despite the fact that FlowNet architecture utilizes larger kernel sizes that the original UNet, presumably to increase the receptive fields in the fine scale pathways. This observation is confirmed by testing a 2D version of the VoxelMorph architecture \citep{balakrishnan2018unsupervised}, which is more similar to the original UNet in its use of small $3\times 3$ kernels. The results were even worse than those for FlowNet and were thus omitted. In contrast, recursive architectures like SPyNet and SEAMLeSS do not suffer from this problem because residual displacements for each aligner network.

\textbf {Direct optimization is slow. } Our comparison of optimization during training and during inference time (shown in Fig.\ref{fig:qqopt}) shows that directly optimizing the transformation field for each image pair can give high-quality results, but it suffers from relatively high computational demands as well as unserviceable worst-case performance. Considering Fig \ref{fig:qqopt}, optimization at inference time seems to perform better than SEAMLeSS or Alembic in the average and best case, but fails more catastrophically in the worst case. We examined some of the latter cases, and found that a common failure mode was becoming trapped in a bad local minimum at a high MIP level, and being unable to correct that in lower MIP levels. Once trained, SEAMLeSS is roughly two orders of magnitude faster to align a particular image pair than optimization at inference time. This is preliminary confirmation of our starting hypothesis that SEAMLeSS can rapidly compute an approximate minimum of Eq. (\ref{eq:LossFunctionStage}). We are not sure whether the speed difference is fundamental, or implementation-dependent (PyTorch); further testing is needed.

\section{Conclusion}
We present a method of image alignment based on hierarchical siamese encoding learned in a self-supervised manner. We show that alignment based on such feature encoding outperforms pixel-based image alignment methods. During training, the nonsmoothness penalty is masked out around discontinuities, which enables the network to correct discontinuous image deformations during inference.  We also show that our approach outperforms one-shot approaches to alignment, because the fine pathways in the latter approach may fail to contribute to alignment accuracy when displacements are large.

\section*{Acknowledgments}
We thank T. Macrina for helpful discussions. We thank our colleagues at the Allen Institute for Brain Science (A. Bodor, A. Bleckert, D. Bumbarger, N. M. da Costa, C. Reid) for the ssEM images that were used in this study. This research was supported by the Intelligence Advanced Research Projects Activity (IARPA) via Department of Interior/ Interior
Business Center (DoI/IBC) contract number D16PC0005, NIH/NIMH (U01MH114824), NIH/NINDS (U19NS104648), NIH/NEI (R01EY027036, R01NS104926), and ARO (W911NF-12-1-0594).
The U.S. Government is authorized to reproduce and
distribute reprints for Governmental purposes notwithstanding any copyright annotation thereon.
Disclaimer: The views and conclusions contained herein are those of the authors and should not be interpreted as
necessarily representing the official policies or endorsements, either expressed or implied, of IARPA, DoI/IBC, or the
U.S. Government.  

\bibliographystyle{plainnat}
\bibliography{seamless}
\newpage

\appendix
\section{Appendix}

\subsection{Training data augmentation}\label{augmentation}
\begin{enumerate}
\item \textbf{Translation.} For a SEAMLeSS network with $l$ levels ($(l-1)$ downsamples), we add random translations of $2^{l-1}$ pixels in both the $x$ and $y$ directions to both $S$ and $T$. The size of the translations are selected so that the top residual network in the pyramid (the residual network operating at the lowest resolution) will be responsible for correcting translations of size roughly 2 pixels.
\item \textbf{Rotation.} We apply independently drawn random rotations to both $S$ and $T$ of $r\sim N(0,0.003)$ radians.
\item \textbf{Scaling.} We independently randomly scale each input slice by a factor $f\sim N(1,0.005)$.
\item \textbf{Lighting.} The dynamic range of each slice is randomly compressed by a factor $d\sim U(1,3)$ by either raising the minimum intensity value of the image or lowering the maximum intensity value of the image (and scaling all pixels correspondingly). Real EM data can show drastic variations in lighting and contrast in a relatively small number of samples. Fig.\ref{lighting} contains examples of real lighting variations in Pinky40. When adjusting lighting as an augmentation, we compute MSE by aligning the unadjusted raw slice $S$, rather than the re-lit version of $S$, simulating a ground truth for real lighting variations. 
\begin{figure}[!ht]
	\centering
	\subfigure[Unusually low illumination.]
	{\centering\includegraphics[width=0.25\textwidth]{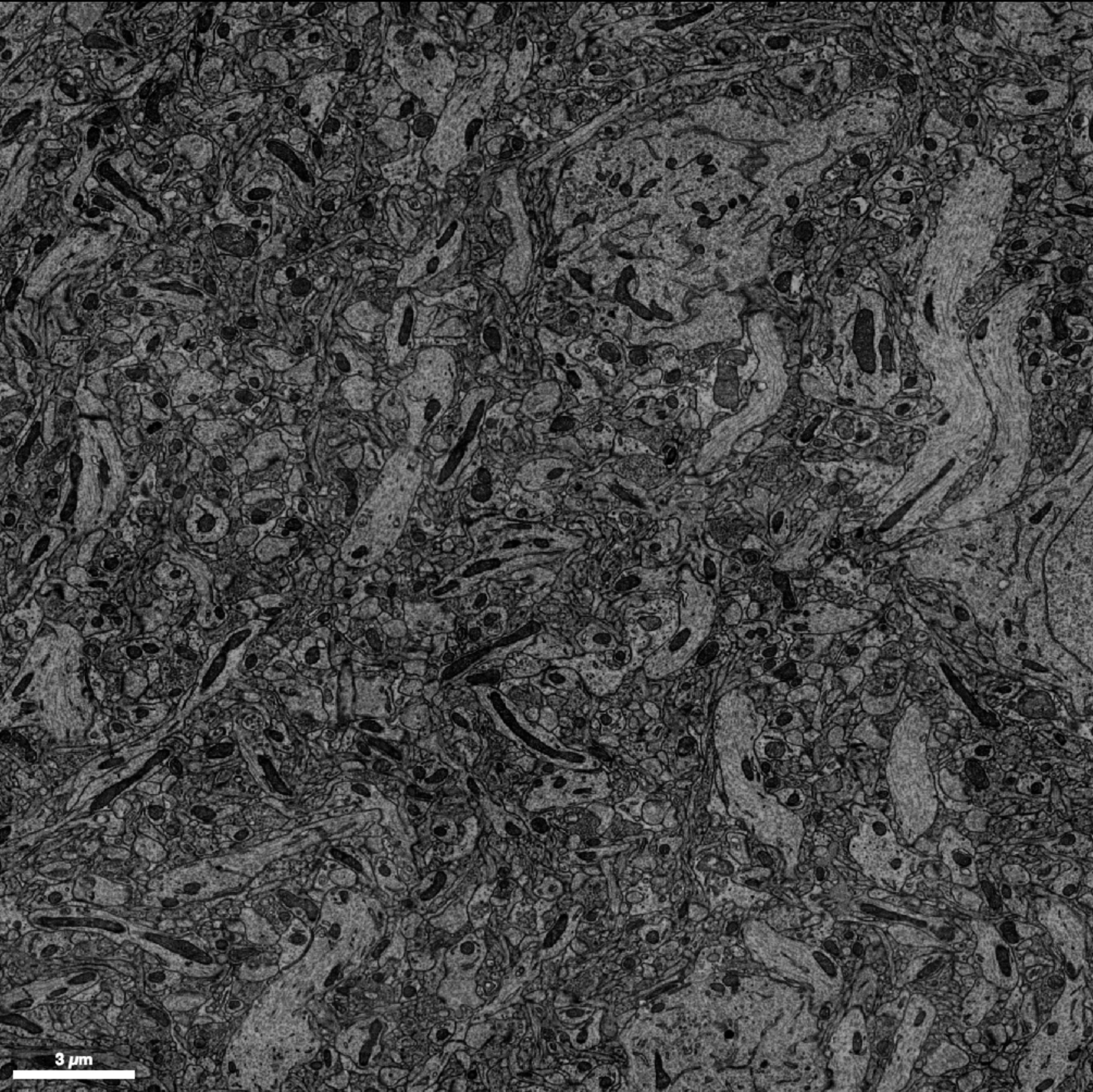}}
	\subfigure[Typical illumination.]
	{\centering\includegraphics[width=0.25\textwidth]{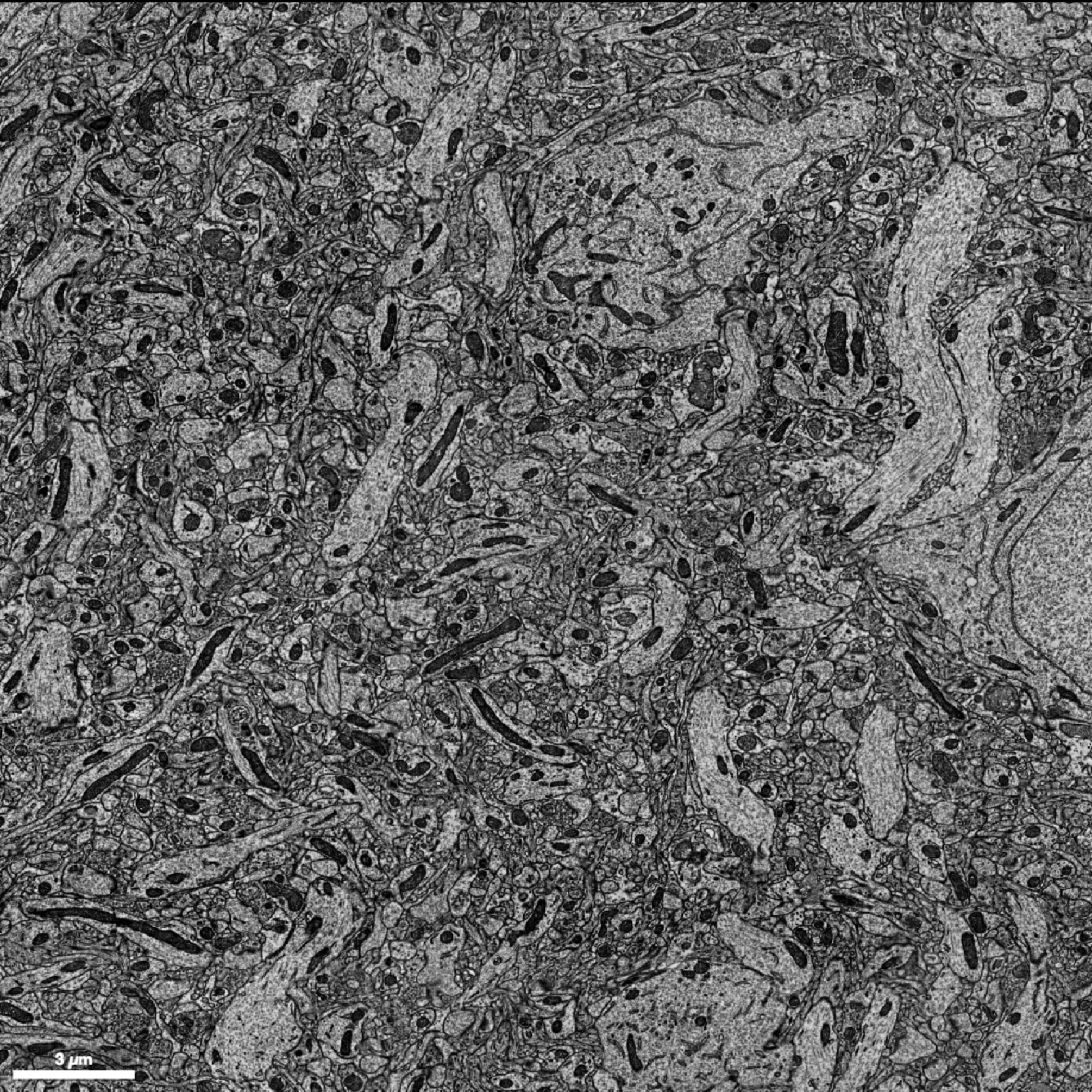}}
	\subfigure[Unusually bright illumination.]
	{\centering\includegraphics[width=0.25\textwidth]{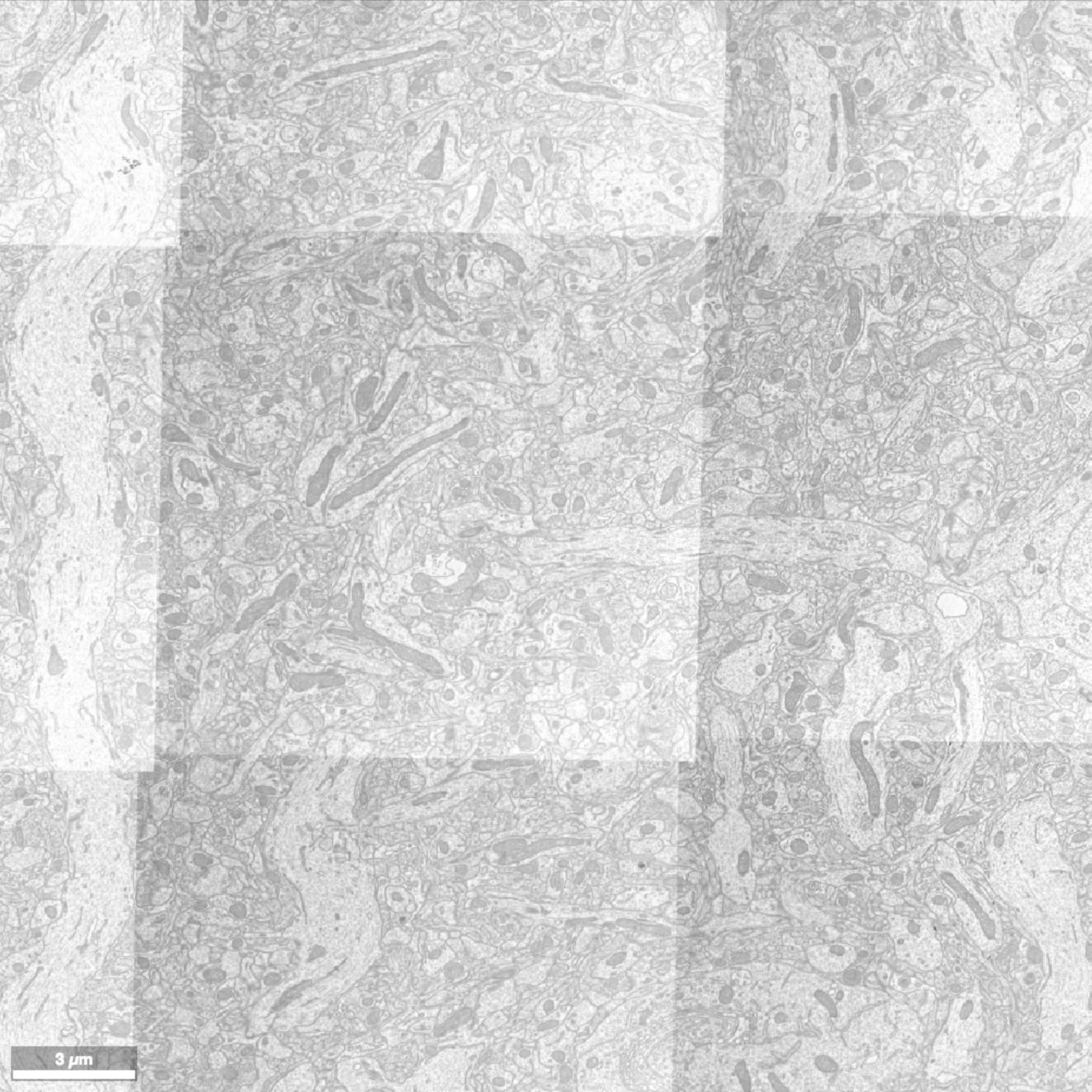}}
	\caption[Various lighting conditions in Pinky40]{Varying lighting conditions in the Pinky40 EM dataset.}\label{lighting}
\end{figure}
\item \textbf{Defects.} We simulate defects (small occluded areas) in the real data by randomly cutting out squares of inputs, setting the corresponding pixel intensities to zero. Ideally, our network should perform as robustly as possible in locations near the defect. An example of a defect in Pinky 40 can be found in Fig.\ref{defect}.
\begin{figure}[!ht]
	\centering
	\includegraphics[width=0.6\textwidth]{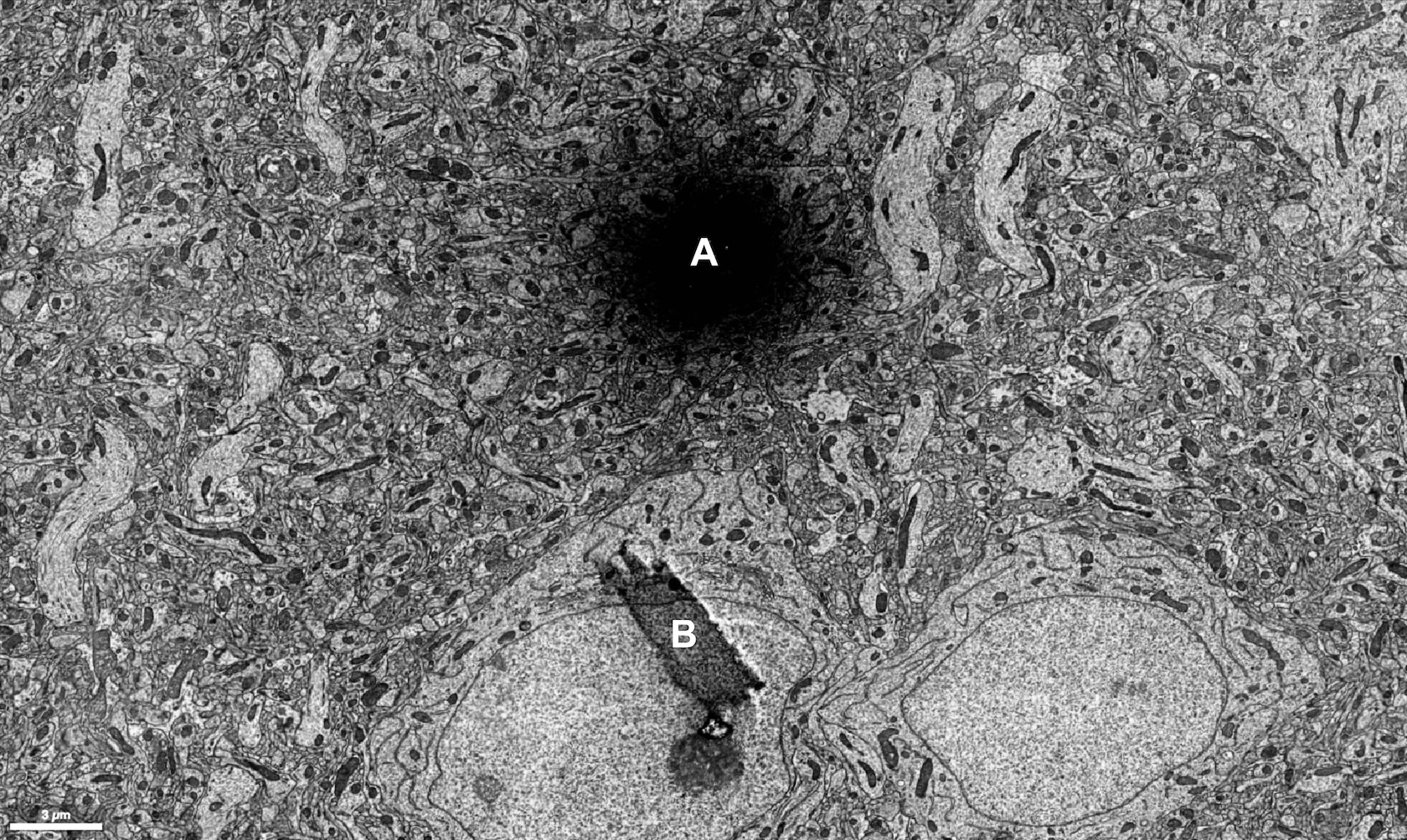}
	\caption[Example of two defects in Pinky40]{Example of two defects in Pinky40, labeled A and B.}\label{defect}
\end{figure}

\item \textbf{Cropping.} In order to prevent the network from using `edges' of the region of valid data to align slices, we randomly slice off edges and corners from inputs such that the correct alignment contains significantly sizes regions where only one of the target slice and the prediction have data.
\end{enumerate}

\subsection{Cracks and Folds}

\subsubsection{Simulating cracks}

When training the SEAMLeSS network to correct discontinuities like cracks, we artificially add cracks to training examples using a simple algorithm. We first generate a seam along which the image will be pulled apart. We then pull the image apart along this seam a randomly generated amount, limited by the assumption that the top aligner network can correct a displacement of one pixel. The random walk used to generate the seam is detailed below:
\begin{enumerate}
    \item Select a random starting pixel in the top row of pixels in the source slice $S$ (pixel (0,$j$) for some random starting point $j$)
    \item Complete a random walk down the image, moving the seam location one pixel right with probability $\frac{p}{2}$, left with probability $\frac{p}{2}$, and moving the seam location directly downwards with probability $(1-p)$. For each crack, $p$ is chosen randomly as $\frac{1}{c}, c\in [3,10]$.
\end{enumerate}

\begin{figure}[htp]
    \centering
    \subfigure[A light simulated crack]
        {\includegraphics[width=0.3\textwidth]{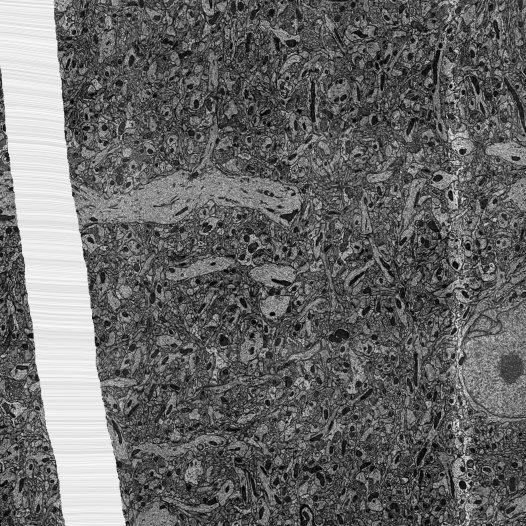}}
    \subfigure[A dark simulated crack]
        {\includegraphics[width=0.3\textwidth]{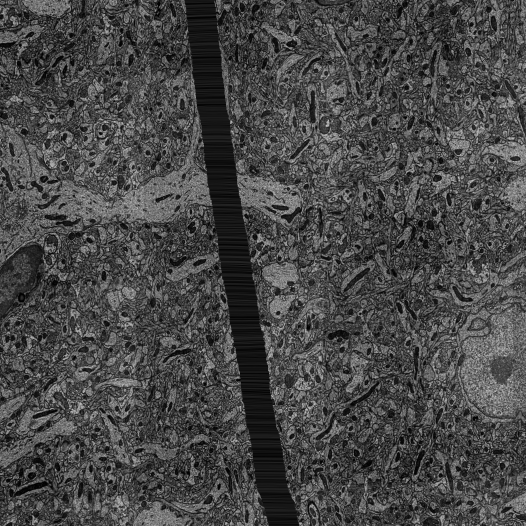}}
    \caption{Two simulated cracks used for training to correct certain non-smooth deformations.}
    \label{simulatedcracks}
\end{figure}

Each simulated crack is also parameterized by a color; half of our cracks are dark, and half are light. When filling in the missing region corresponding to the crack, we use pixel values drawn randomly from $[0.01,0.15$ for dark cracks and $[0.85,0.99]$ for light cracks. Examples of simulated cracks are shown in Fig.\ref{simulatedcracks}. Because input images are rotated randomly during training, this process effectively generates cracks of every orientation.

\subsubsection{Examples}
We include examples of real cracks in folds in our EM datasets.

\begin{figure}[htp]
	\centering
	\subfigure[The image slice before the crack.]
	{\centering\includegraphics[width=0.3\textwidth]{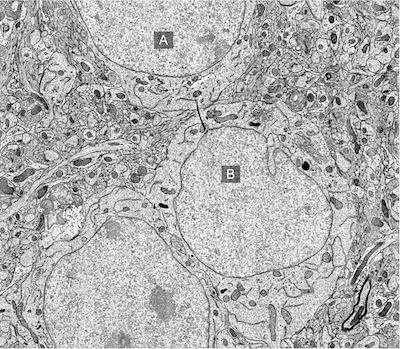}}
	\subfigure[The image slice containing the crack.]
	{\centering\includegraphics[width=0.3\textwidth]{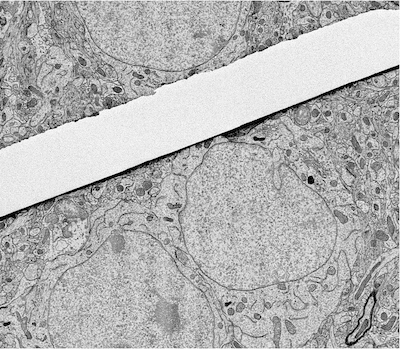}}
	\caption[Example of a crack in the Basil EM dataset]{An example of a crack from another dataset, called Basil. We show two consecutive slices, $k,k+1$, where slice $k+1$ contains the crack. Compare the distance between the cell bodies labeled \textbf{A} and \textbf{B} in the top-right region of (a) with their separation in (b). While cracks of this size are less common in Pinky40, they are still a problem.}\label{cracks}
\end{figure}

\subsection{Discontinuities can be learned by masking out the nonsmoothness penalty}

We show compelling quantitative and qualitative evidence not only for the effectiveness of masking for learning smooth solutions with allowance for singularities (Figs.~\ref{crackmaskcompare} and~\ref{fig:crackqq}), but also for the ineffectivness of naively training on non-smooth samples without masking (Fig.~\ref{fig:qqcracktrainnnocrack}). The technique of masking regions where the solution is known to be discontinuous can be generally applied to various types of non-smooth deformations, whether through augmentation or detection by another method.

While we did not make this comparison explicitly in this work, we suspect that using learned feature maps instead of raw downsampled pixels might allow for better performance in the face of non-smooth deformations, as learned features maps likely provide denser information to each aligner net. Crucially, this should allow alignment decisions to be made more locally, without consensus from a very wide region (which is necessary in the case of raw downsampling, as only very large structures remain at high MIP levels). More local alignment decisions intuitively allow for more accurate correction of non-smooth deformations.

\begin{figure}[htp]
	\centering
	\subfigure[The image slice before the fold.]
	{\centering\includegraphics[width=0.3\textwidth]{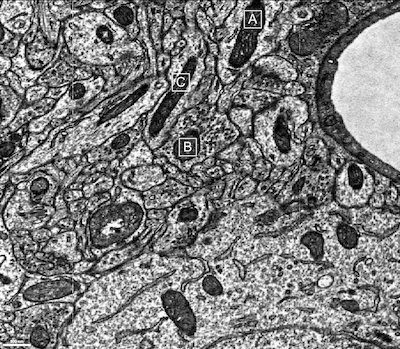}}
	\subfigure[The image slice containing the fold.]
	{\centering\includegraphics[width=0.3\textwidth]{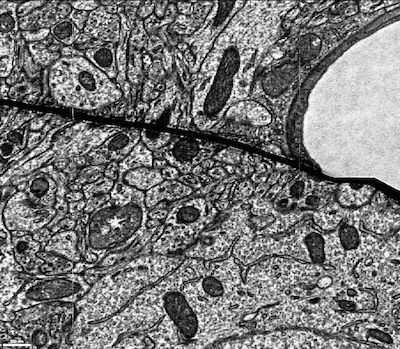}}
	\caption[Example of a fold in the Pinky40 EM dataset]{An example of a fold in the Pinky40 dataset. We show three consecutive slices, $k,(k+1)$, where slice $k+1$ contains the fold. Compare the distance between objects labeled \textbf{A} and \textbf{B} in the top-central region of (a) with their separation in (b). Note also the contraction of object \textbf{C} in (b).}\label{folds}
\end{figure}

\subsection{Connectomics: very large-scale inference}
In applying this work to the field of Connectomics, the presented model needs to be applied at extremely large scale in order to align super-resolution electron microscopy stacks. The challenge lies in the size of the images: an image corresponding to a $0.1\times 0.1$mm slice of tissue has a resolution of roughly $55000\times 34000$ pixels. For a neural network architecture with even a very modest number of feature maps, this is far too large to fit into a GPU. This means that the inference pipeline must process images in a chunked manner, while producing seamless results at chunk boundaries.  

Given an unaligned stack, the inference pipeline will align image pairs consequently, starting with the first image as a reference. This carries the implicit assumption that this first image is not deformed. From the inference perspective, alignment of two images consists of two stages -- field generation and image warping. In the field generation stage the model is applied to produce correspondence fields between source and target images. In the image warping stage the source image is warped according to the produced field. 

For the field generation stage, the source image is broken into non-overlapping chunks. The atomic task for this stage is producing the dense vector field of correspondences for a single source image chunk. If the network were naively applied to the source image chunks, the resulting vector fields would be inconsistent at the boundaries due to zero padding. In order to have consistent, seamless results on chunk boundaries, the network outputs have to be cropped. Cropping the output by the field of view of the network guarantees perfect consistency, but in practice cropping beyond $256$ pixels does not improve the output quality.

Because each chunk has to be cropped, increasing the chunk size relative to the crop amount reduces the proportion of wasted computation, although the chunk size is strictly limited by the amount of memory available on the system.

Analogously, in the image warping stage the source image is also broken into non-overlapping chunks. The chunk size for the image wrapping stage is generally much larger than for the field computation stage, as during the image warping stage, only the pre-warped image, post-warped image, and corresponding warp field have to be stored in memory. On the other hand, during the field computation stage, not only the image but also the intermediate activation values have to be stored in memory while SEAMLeSS is being applied. 

The atomic task in the image warping stage is producing the warped output for the given chunk of the destination image. Theoretically, the whole source image could be wrapped into a single chunk in the destination image, requiring the whole source image to be loaded into memory for chunk processing. In practice however, it is possible to put limitations on the maximum displacements produced by the network based on the dataset being processed. For pre-aligned EM images, the displacements range from $2000$ to $5000$ pixels. To produce a post-warped chunk in the source image the worker has to load the region of the pre-warped that exceeds the chunk by the maximum displacement value.

\end{document}